\documentclass[]{spie}  

\usepackage{amsmath,amsfonts,amssymb}
\usepackage{graphicx}
\usepackage[colorlinks=true, allcolors=blue]{hyperref}
\usepackage{multirow}
\usepackage{float}
\newcommand{\bftab}{\fontseries{b}\selectfont}

\title{Unsupervised Synthetic Image Refinement via Contrastive Learning
and Consistent Semantic-Structural Constraints}

\author[a]{Ganning Zhao}
\author[b]{Tingwei Shen}
\author[c]{Suya You}
\author[a]{C.-C. Jay Kuo}
\affil[a]{University of Southern California, Los Angeles, CA, USA}
\affil[b]{University of California, Berkeley, Berkeley, CA, USA}
\affil[c]{DEVCOM Army Research Laboratory, Los Angeles, CA, USA}

\authorinfo{Further author information: (Send correspondence to Ganning Zhao)\\
Ganning Zhao: E-mail: ganningz@usc.edu}

\begin{document} 
\maketitle

\begin{abstract}

Ensuring the realism of computer-generated synthetic images is crucial
to deep neural network (DNN) training.  Due to different semantic
distributions between synthetic and real-world captured datasets, there
exists semantic mismatch between synthetic and refined images, which in
turn results in the semantic distortion.  Recently, contrastive learning
(CL) has been successfully used to pull correlated patches together and
push uncorrelated ones apart. In this work, we exploit semantic and
structural consistency between synthetic and refined images and adopt CL
to reduce the semantic distortion. Besides, we incorporate hard negative
mining to improve the performance furthermore.  We compare the
performance of our method with several other benchmarking methods using
qualitative and quantitative measures and show that our method offers
the state-of-the-art performance. 

\end{abstract}


\section{INTRODUCTION}\label{sec:intro}  

Synthetic image refinement, aiming at making synthetic images more
realistic to human observers, is an increasingly important research
topic due to its numerous applications. Machine learning models with a
huge number of parameters (e.g., deep neural networks) demand vast
amounts of training data to ensure good performance.  However,
real-world data acquisition, including data collection and labeling, is
expensive and time-consuming. To overcome this challenge, researchers
have turned to computer-generated synthetic data as a substitute for
real-world data during training. Nevertheless, there exists a
significant gap between the distributions of real and synthetic data.
This gap tends to degrade the model performance in downstream tasks. 

To narrow the gap, one solution is to improve the generator so that it
can generate more realistically-looking synthetic data.  Generating
realistic synthetic data remains challenging and costly, as a powerful
generator cannot produce close-to-real synthetic data.  Another solution is to learn the
characteristics from images captured in real world and improve the
realism of synthetic images accordingly.  In this work, we take the
second path.  Simply speaking, we refine the synthetic data by improving
their realism with an objective that they are similar to the real data
in distributions. 

Various lines of research have been proposed to refine synthetic images\cite{shen2023study}.
As one of the first papers in this field, SimGAN
\cite{shrivastava2017learning} used an adversarial structure with
limited success. Many image-to-image translation methods are proposed to
transfer input images in the source domain to output ones in the target
domain. DiscoGAN \cite{kim2017learning}, CycleGAN
\cite{zhu2017unpaired}, and DualGAN \cite{yi2017dualgan} applied a
bidirectional structure to ensure cycle-consistency between source and
target images.  However, the assumption of bijective mapping may not
hold in practice since images from source domain may not have
counterparts in target domain, and vice versa. To address this issue,
various one-sided image translation methods have been proposed,
including DistanceGAN \cite{benaim2017one} and CcGAN
\cite{fu2019geometry}, However, the semantic distortion problem still
exists. There is semantic mismatch between input and output images. 

Recently, contrastive learning (CL) has been adopted for image-to-image
translation. One pioneering work is CUT \cite{park2020contrastive}.
Afterward, researchers came up with various ideas for further
improvement such as deriving more powerful loss functions
\cite{yeh2022decoupled, chen2021simpler}, utilizing more informative
negative samples \cite{wang2021instance, hu2021adco,
robinson2020contrastive}, and combining these two ideas.  However, most
of them have not been tailored to synthetic images the refinement.
Inspired by the recent success of CL in image-to-image translation, we
propose an unsupervised CL method to improve the realism of synthetic
images and reduce the semantic distortion. 

It was pointed out in \cite{guo2022alleviating} that the semantic
distortion is caused by the difference in semantic distributions of
source and target domains. For instance, consider the refinement of
synthetic images in the GTA5 dataset to the real-world CityScapes
domain. There are more vegetation and cars in CityScapes than those in
GTA5. For example, trees appear in the sky and building regions and cars
appear in the street in CityScapes. The difference causes semantic
inconsistency between input synthetic and output refined images,
resulting in downstream semantic segmentation errors. Furthermore, the
structure information is highly correlated with the semantics. For
instance, buildings, trees, and roads have different structures. To
reduce the semantic distortion, we incorporate both semantic and
structural information in CL to ensure semantic consistency between
input and output images. 

This work has three main contributions.
\begin{enumerate}
\item We proposes a novel CL method that can enhance the realism of
synthetic images. Our method provides additional training data to
improve downstream tasks and has the potential to aid in computer game
rendering. 
\item We derive a semantic-structural relation consistency loss (SSRC) to
maintain semantic and structural consistency between input synthetic and
output refined images.
\item We conduct experiments on various datasets to demonstrate the
effectiveness and generalizability of our method in improving the
realism of synthetic images.  Our method offers state-of-the-art
performance through quantitative and qualitative evaluations. 
\end{enumerate}

\section{Related Work}\label{sec:Related Work}

\subsection{Early Work on Synthetic Image Refinement}

SimGAN \cite{shrivastava2017learning} was one of the pioneering papers
that improve the realism of synthetic images using unlabeled real data,
while preserving their annotation information with an adversarial
network. It added an L1 self-regularization loss in the image pixel
space to keep the annotations. Based on this method, Atapattu et al.
\cite{atapattu2019improving} proposed an improved version that
incorporated a perceptual loss term to further enhance the realism of
the refined images. However, their methods only applied to gray-scale
images with simple structures. Their performance in complex-scene
datasets is limited. 

CyclGAN \cite{zhu2017unpaired}, DualGAN \cite{yi2017dualgan}, and
DiscoGAN \cite{kim2017learning} assumed a bijective translation function
and used the cycle consistency loss to enforce the consistency between
images in the source domain, $X$, and the reconstructed domain,
$F(G_{XY}(X))$, and vice versa. In this setting, $G$ is the mapping from
the source domain, $X$, to the target domain, $Y$, and $F$ is the
inverse mapping from the target domain, $Y$, to the source domain, $X$.
These methods require an additional generator/discriminator pair.
Additionally, they are restricted to bijective image translation.
\cite{lee2018diverse, park2020contrastive, wang2021instance}.  As a
result, they may work for synthetic image refinement since the source
and target domains may not contain the same scenes or content. 

One-sided image translation methods were proposed as alternatives to the
cycle-consistent constraint. DistanceGAN \cite{benaim2017one} learns the
one-sided mapping by ensuring a strong correlation between pairwise
distances calculated in the source and target domain. CcGAN
\cite{fu2019geometry} predefines a geometric transformation $f(.)$ so
that $f^{-1}(f(x))=f(f^{-1}(x))=x$ and enforces geometry-consistency
between an input image and its counterpart transformed by $f(x)$ in the
new domain (i.e. $f(G_{XY}(x))\approx G_{XY}(f(x))$). 

\subsection{Contrast Learning (CL) and InfoNCE Loss}

A few CL methods have been proposed to improve image translation tasks
recently. CUT \cite{park2020contrastive,zheng2021spatially} utilized
mutual information maximization to learn a cross-domain similarity
between patches from the source and the target domains.  NEGCUT
\cite{wang2021instance} trained a generator to produce negative samples
online and replace the random negative samples, which effectively
brought the positive and query samples closer. F-LSeSim
\cite{zheng2021spatially} minimized fixed self-similarity and learned
self-similarity to keep scene structures. The first loss is computed as
a spatially-correlative map between input and output patches, and the
second loss is computed as the contrastive loss among input, augmented
input and target domain patches. 

Several loss functions have been proposed to improve th CL model
performance. One famous one is the InfoNCE loss
\cite{oord2018representation}, which describes the relation of
correlated (positive) samples over both correlated and uncorrelated
(negative) samples. Minimizing this loss function, while maximizing the
lower bound of their mutual information, makes the feature space of
positive samples more correlated and that of negative samples less
correlated. This technique and the corresponding loss function have been
popular, e.g., \cite{bachman2019learning, park2020contrastive,
chen2020simple, he2020momentum}. 

Variants of InfoNCE have been investigated to further improve the
performance. To give an example, the negative-positive coupling (NPC)
effect reduces learning efficiency due to vanishing gradients caused by
the loss function. This occurs when randomly selected negative patches
are uncorrelated with positive ones. The decoupled infoNCE loss function
\cite{yeh2022decoupled} can address this problem by removing the
positive pair term in the denominator.  Another idea to improve the
performance is mining hard negative samples, e.g., using adversarial
training on negative samples \cite{hu2021adco} or applying a more
informative negative sampling strategy based on the von Mises Fisher
distribution \cite{robinson2020contrastive}. 

\begin{figure} [ht]
\begin{center} \begin{tabular}{c}
\includegraphics[height=10cm]{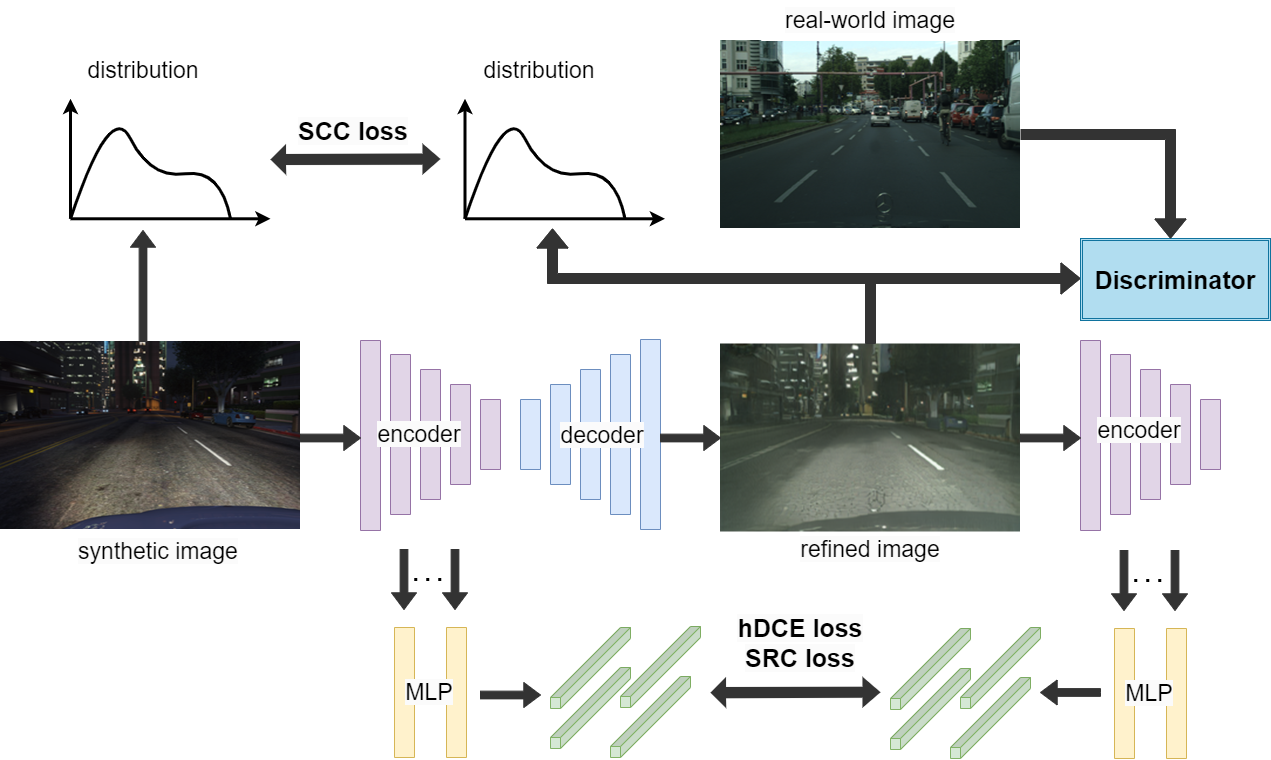}
\end{tabular} 
\end{center}
\caption[1]{System overview of the proposed method.}\label{fig:pipeline} 
\end{figure} 

\section{Proposed Method}\label{sec:Methodology}

\subsection{System Overview}\label{sec:method_overview}

\begin{figure} [ht]
\begin{center} \begin{tabular}{c}
\includegraphics[height=4.5cm]{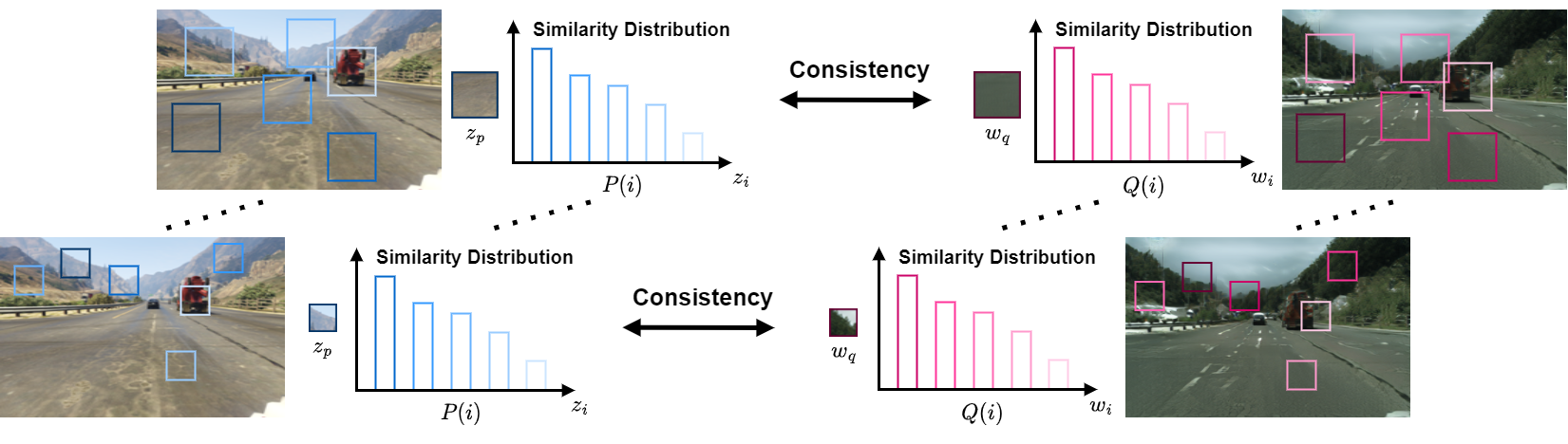}
\end{tabular} \end{center}
\caption[1]{Illustration of the SRC loss. The encoder extracts features
from different layers, which correspond to different patch sizes. We use
pseudo labels to represent the similarity distribution, assigning higher
pseudo scores to patches with closer semantic relations to a given query
or positive patches.} \label{fig:SRC}
\end{figure}

We aim to design a CL-based method to refine synthetic images so that
they appear more realistic to humans. The system consists of three parts.
\begin{itemize}
\item As shown in the middle row of Figure \ref{fig:pipeline}, input
synthetic images are fed into a generator with an encoder/decoder
structure to generate refined output images.  Afterward, the refined
images are fed back into the encoder with the same weights. 
\item As shown in the bottom row of Figure \ref{fig:SRC}, the encoded
features from different layers of synthetic and refined images are fed
into a 2-layer MLP to yield a shared embedding space, denoted by
$\mathcal{H}$. Since deeper layers in the encoder have larger receptive
fields, the spatial positions of deeper features correspond to larger
patches in synthetic images. Hence, different patches correspond to
different feature vectors in the MLP feature space, and deeper layers'
features correspond to larger patches, Next, we propose the
semantic-structural relation consistency (SSRC) loss to constrain
semantic and structural distortions.  Furthermore, we use the hard
negative decoupled infoNCE loss (hDCE) to sample hard negative patches. 
\item As shown in the top row of Figure \ref{fig:pipeline}, as synthetic
images learn attributes from real-world captured images, the real-world
images and refined images are fed to a discriminator to reduce their
differences. 

\end{itemize}

We adopt the following notation for the rest of this section.  Let
synthetic images $X \in \mathcal{X}$, refined images $\hat{Y} \in
\mathcal{\hat{Y}}$ and real-world images $Y \in \mathcal{Y}$, where
$\mathcal{X}$, $\mathcal{Y}$ and $\mathcal{\hat{Y}}$ are the source,
target, and destination domains. Patches in their correspondings images
are represented by $x \in X$, $y \in Y$ and $\hat{y} \in \hat{Y}$.
Embedding vectors from MLP feature space of different patches are $z \in
x$, $w \in y$ and $\hat{w} \in \hat{y}$. 

\subsection{Semantic-Structural Relation Consistency (SSRC)}\label{sec:SSRC}

The semantic-structural relation consistency (SSRC) loss contains
two complementary components: 1) semantic relation consistency (SRC)
and 2) structure consistency constraint (SCC). The SRC loss, $L_{SRC}$,
is computed in the feature space to alleviate the semantic distortion, while
the SCC loss, $L_{SCC}$, is computed in the pixel space to reduce the structural
distortion. The SSRC loss is given by the weighted sum of $L_{SRC}$ and $L_{SCC}$:
\begin{equation}\label{eq:SSRC_loss}
L_{SSRC} = \lambda_{SRC}L_{SRC} + \lambda_{SCC}L_{SCC}.
\end{equation}
We will derive $L_{SRC}$ and $L_{SCC}$ below.

\subsubsection{Semantic Relation Consistency (SRC)}\label{sec:SRC}

Most CL losses use a one-hot labeling system to differentiate patches. A
label of one is assigned to positive patches and a label of zero to
negative ones. However, this approach is uninformative because it treats
all negative samples equally, while their semantic similarity to the
query patch varies and thus contributes to the loss with different
weights. To address this issue, we encourage consistency between the
semantic relation of the query and negative patches in the output image,
and that of positive and negative patches in the input image. 

We capture the semantic relation by soft pseudo labels
\cite{wei2020co2,yang2022mutual,jung2022exploring} of negative patches,
which reveal more information about their correlation with a query or
positive patches. Specifically, for an output image, we define the
semantic similarity distribution between a query, $x_q$, and a negative
patch, $x_i^-$, in the feature space as a softmax function:
\begin{equation}
Q(i) = \frac{\exp{z_q^Tz_i^-}}{\sum_{j=1}^{K} \exp{z_q^Tz_j^-}},
\end{equation}
where $z_q$ and $z_i^-$ are the embedding vectors of $x_q$ and $x_i^-$,
respectively. Similarly, for an input image, the semantic similarity between a
positive patch, $y_p$, and a negative patch, $y_i^-$, is defined as
\begin{equation}
P(i) = \frac{\exp{w_p^Tw_i^-}}{\sum_{j=1}^{K} \exp{w_p^Tw_j^-}},
\end{equation}
where $w_p$ and $w_i^-$ are the corresponding embedding vectors of $y_p$
and $y_i^-$, respectively.  We apply the Jensen-Shannon Divergence (JSD)
to preserve the consistency between the probability distribution of
$P_k$ and $Q_k$ as shown in Figure \ref{fig:SRC}. Then, we have
\begin{equation} \label{eq:L_SRC}
L_{SRC} = \sum_{k=1}^{K} \widehat{JSD} (P_k || Q_k) 
= \frac{1}{2} KL(P||M) + \frac{1}{2} KL(Q||M),
\end{equation}
where $M = \frac{1}{2} (P+Q)$. Therefore, the minimization of
Eq. (\ref{eq:L_SRC}) will constrain the semantic consistency between 
input and output patches.

\subsubsection{Structure Consistency Constraint (SCC)}\label{sec:SCC}

Previous studies \cite{jia2021semantically, guo2022alleviating}
illustrate significant differences in class statistics between datasets.
For instance, more pixels are labeled as ``sky" and ``building" while
fewer are labeled as ``vegetation" in the GTA5 dataset as compared to
those in the CityScapes dataset. When we use GTA5 as the source and
CityScapes as the destination. Sky or building regions in GTA5 images
are more prone to be falsely transformed into vegetation in refined
target images. This is called the structure distortion between the source
and target images. It degrades the refinement of synthetic data because
of the semantic distortion. 

To address this issue, we develop a new loss term called the structural
consistency loss. It operates in the image pixel space to minimize the
structure distortion. More structure information can be preserved if the
corresponding pixels in input and output images have stronger
correlation.  Intuitively, a realistic style should be added to the
input images, instead of changing the image shape.  Thus, maximizing the
mutual information between input and output pixels enforces structural
consistency, as it amplifies non-linear dependencies between them.  To
achieve efficient and numerically stable computation, we adopt the
relative Squared-loss Mutual Information (rSMI), which is an extension
of the Squared-loss Mutual Information \cite{guo2022alleviating}. 

For an input image of the source domain, $X_i \in \mathcal{X}$, and an
output image of the target domain, $\hat{Y_i} \in \mathcal{Y}$, we use
$V^{X_i}$ and $V^{\hat{Y_i}}$ to indicate random variables for pixels in
$X_i$ and $\hat{Y_i}$. The structure consistency constraint (SCC)
is defined as
\begin{equation}
L_{SCC} = - \frac{1}{N} \sum_{i=1}^{N} \widehat{rSMI} (V^{X_i}, V^{\hat{Y_i}}),
\end{equation}
where $N$ is the sample number, and $rSMI$ is computed by the relative Pearson(rPE) 
Divergence \cite{yamada2013relative} via
\begin{equation}
rSMI(V^{X_i}, V^{\hat{Y_i}}) = D_{rPE}(P_{V^{X_i}} \otimes V^{\hat{Y_i}} || 
(P_{(V^{X_i}, V^{\hat{Y_i}})}).
\end{equation}

\subsection{Hard Negative Decoupled infoNCE (hDCE)}\label{sec:hDCE}

If the negative patches are uncorrelated with the query patch (i.e.,
easy negative samples), the gradient of InfoNCE will be small due to a
large denominator, which potentially causes gradient vanishing and
affects learning efficiency. This is known as negative-positive coupling
(NPC). It is desired to mine negative patches that are semantically
correlated with the query patch (i.e., hard negative samples) and use a
different loss function to alleviate the NPC effect. We re-sample
negative samples with the von MisesFisher distribution
\cite{robinson2020contrastive} to obtain different latent classes and 
achieve large semantic similarity (inner product) between query and negative
patches. Mathematically, we have
\begin{equation}
z^- \sim q_\beta (z^-), \quad \textrm{where} \quad q_\beta (z^-) 
\varpropto e^{\beta z^T z^-} \cdot p(z^-),
\end{equation}
where parameter $\beta$ is a concentration parameter that controls the
similarity of hard negative samples with query samples. 

As discussed in Section \ref{sec:Related Work}, decoupled infoNCE (DCE)
alleviates the NPC effect by removing the positive pair term in the
denominator of InfoNCE \cite{yeh2022decoupled}. With the negative
sampling strategy \cite{jung2022exploring}, we can finetune the loss
function furthermore as
\begin{equation}\label{eq:hDCE_loss}
L_{hDCE} = \mathbb{E}_{(z,w)} \left[-log \frac{\exp(w^Tz/\tau)}
{N\mathbb{E}_{z^- \sim q_\beta}[\exp(w^Tz^-/\tau)]} \right],
\end{equation}
where $N$ is the number of negative patches and $\tau$ is the temperature
parameter that controls the strength of penalties on hard negative
samples.

\subsection{Full Objective Function}\label{sec:FullObject}

To ensure the distribution similarity between refined images in the
target domain and real-world images in the destination domain, 
we include the standard adversarial loss in the loss function to obtain
the ultimate loss. The standard adversarial loss is
\begin{equation}\label{eq:GAN_loss}
L_{GAN} (G,D,X,Y) = \mathbb{E}_{y \sim Y}logD(y) + 
\mathbb{E}_{x \sim X}log(1-D(\hat{y})).
\end{equation}
The ultimate full objective fundtion is in form of
\begin{equation}
L_{Full} = L_{SSRC} + \lambda_{hDCE} L_{hDCE} + \lambda_{GAN} L_{GAN},
\end{equation}
where $L_{SSRC}$, $L_{hDCE}$, and $L_{GAN}$ are given in Eqs.
(\ref{eq:SSRC_loss}), (\ref{eq:hDCE_loss}), and (\ref{eq:GAN_loss}),
respectively, and $\lambda_{hDCE}$ and $\lambda_{GAN}$ are weights.

\section{Experiments}\label{sec:Experiments}

We evaluate the performance of our method quantitatively and
qualitatively in this section. We refine the synthetic GTA5 dataset
\cite{richter2016playing} and a more challenging Synthia
(SYNTHIA-RAND-CITYSCAPES) \cite{ros2016synthia} dataset to the domain of
the CityScapes dataset \cite{cordts2016cityscapes} as their labels are
compatible. Specifically, we resize images in all datasets to the size
of $256 \times 256$ and use the standard training/test split in GTA5 and
CityScapes datasets. As no standard training/test split is provided for
Synthia, we use the last 1,880 images for testing and the remaining
7,520 images for training. 

\begin{table}[htbp]
\caption{Quantitative evaluation on our method and benchmarking methods
in \cite{guo2022alleviating}.} \label{tab:evaluations}
\centering
\begin{tabular}{|c||ccc||ccc|}\hline
\multirow{2}{*}{Methods}& \multicolumn{3}{c||}{GTA5 $\rightarrow$ Cityscapes} 
& \multicolumn{3}{c|}{Cityscapes parsing $\rightarrow$ Image} \\ \cline{2-7}
& pixel acc $\uparrow$ & class acc $\uparrow$ & mean IoU $\uparrow$ & 
pixel acc $\uparrow$ & class acc $\uparrow$ & mean IoU $\uparrow$ \\ \hline
GAN+VGG & 0.216 & 0.098 & 0.041 & 0.551 & 0.199 & 0.133 \\ \hline
DRIT++ & 0.423 & 0.138 & 0.071 & \textbackslash & \textbackslash & \textbackslash \\ \hline
GAN & 0.382 & 0.137 & 0.068 & 0.437 & 0.161 & 0.098 \\ \hline
CycleGAN & 0.232 & 0.127 & 0.043 & 0.520 & 0.170 & 0.110 \\ \hline
CUT & 0.546 & 0.165 & 0.095 & 0.695 & 0.259 & 0.178 \\ \hline
CUT+SCC & 0.572 & 0.185 & 0.110 & 0.699 & 0.263 & 0.182 \\ \hline
\bftab Ours & \bftab 0.654 & \bftab 0.186 & \bftab 0.113 & \bftab 0.714 & \bftab 0.263 & \bftab 0.184 \\ \hline
\end{tabular}
\end{table}

\begin{figure} [ht]
\begin{center} \begin{tabular}{c}
\includegraphics[height=9.2cm]{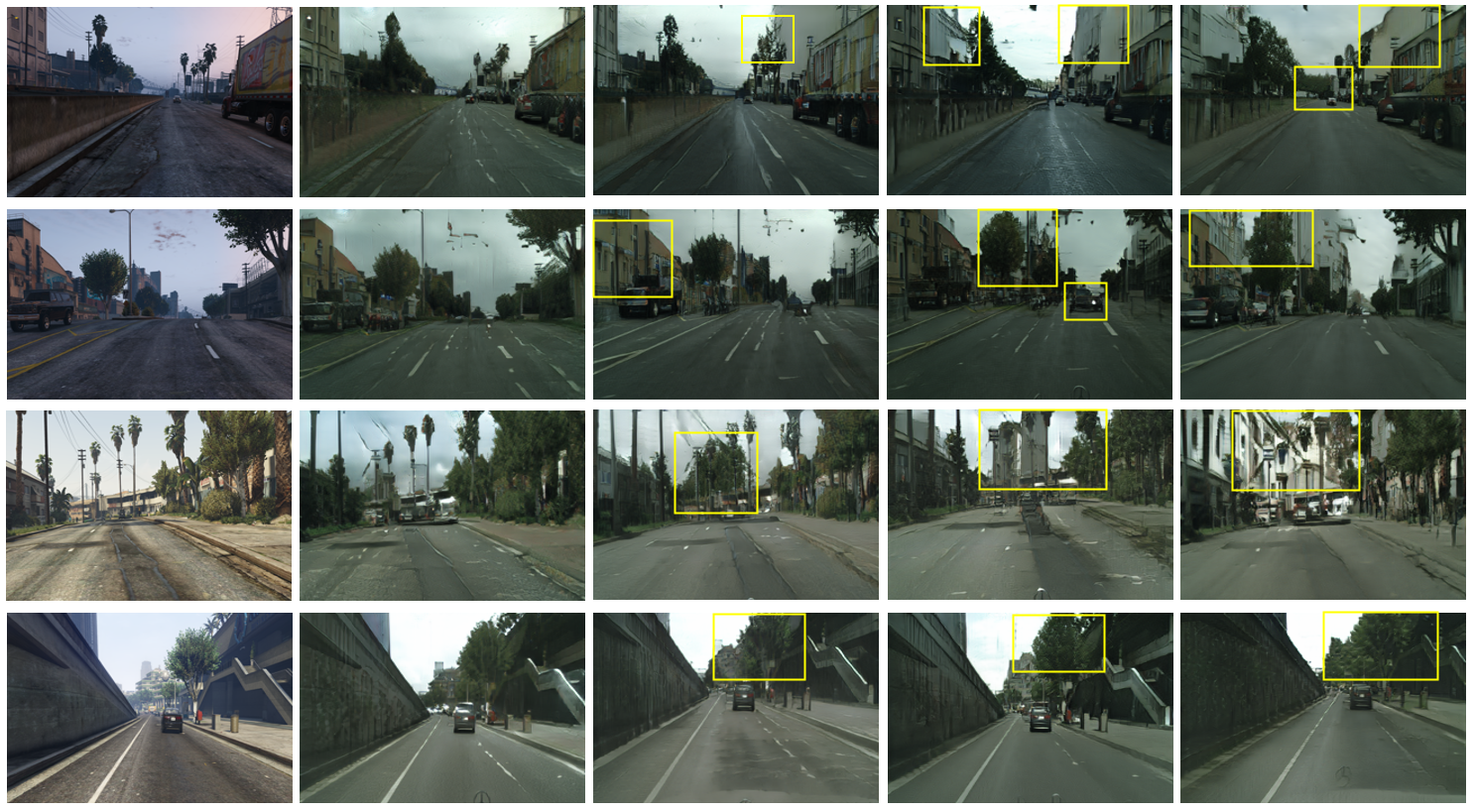} \\
\hspace{0.5cm} Input \hspace{2.2cm} \textbf{Ours} \hspace{2.1cm} CUT \hspace{2.1cm} 
CUT+SCC \hspace{1.9cm} CUT+SRC 
\end{tabular} \end{center}
\caption[1]{Qualitative visual comparison of images refined by our
method and other benchmarking methods.} \label{fig:gta2cs}
\end{figure} 

\begin{figure} [ht]
\begin{center} \begin{tabular}{c}
\includegraphics[height=4.5cm]{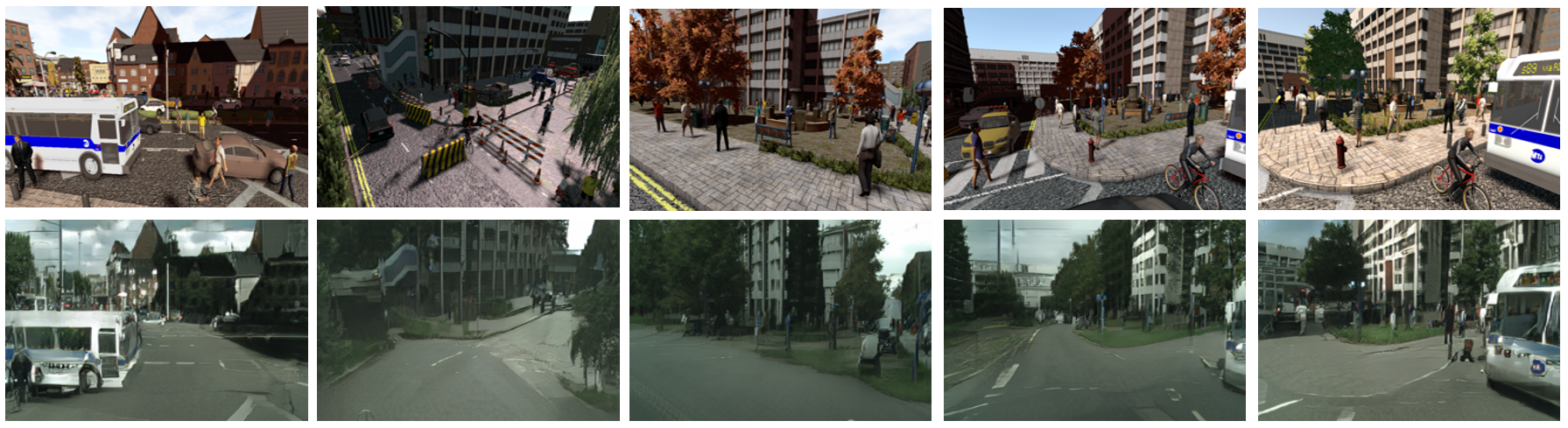} \\
\end{tabular} \end{center}
\caption[1] {Qualitative visual comparison of five images refined by
our method taken from the more challenging Synthia dataset, where the
first row shows five input synthetic images and the second row displays
the correspondent refined images.} \label{fig:syn2cs}
\end{figure} 

\subsection{Quantitative Evaluation}

To quantitatively evaluate the performance of our method in refining
synthetic images and its effectiveness in improving downstream semantic
segmentation tasks, we train a segmentation network on the CityScapes
dataset and then feed the refined synthetic data to it to predict the
segmentation label maps. Then, we compute the accuracy and Intersection
over Union (IoU) scores by comparing the predicted maps with
ground-truth label maps of the synthetic dataset. High scores indicate
its semantic consistency between input and output and thus benefit to
segmentation tasks.  

To be consistent with previous work in \cite{guo2022alleviating}, we
refine the first 500 images of the test set in the GTA5 dataset and feed
them to FCN-8s \cite{long2015fully} as the evaluation network
pre-trained on CityScapes by pix2pix \cite{isola2017image}. The network
prediction is the segmentation label maps, and we compute the scores
with groundtruth label map of GTA5. We compare the performance of our
method with other methods in Table \ref{tab:evaluations}. As shown in
the column ``GTA5 $\rightarrow$ Cityscapes" of the table, our method
offers the state-of-the-art performance with a substantial performance
gap. 

To further demonstrate the effectiveness of our method, we evaluate the
domain mappers in CityScapes utilizing FCN scores. Specifically, we feed
the ground truth segmentation label maps to our model to generate
predicted images. Then, we use the generated images as the input to
FCN-8s to predict the segmentation label maps. We compute the evaluation
metrics between the predicted and ground truth labels. As shown in the
column ``Cityscapes parsing $\rightarrow$ Image" of Table
\ref{tab:evaluations}, our method achieves state-of-the-art performance
in this evaluation. 

\subsection{Qualitative Evaluation}

Figure \ref{fig:gta2cs} shows four representative testing images generated by
four different methods, including Ours, CUT, CUT with the SCC loss only,
and CUT with the SRC loss only. Our method offers superior visual
results by preserving both semantic and structural consistency between
input and output. In contrast, the other three benchmarking methods
exhibit more semantic distortions, such as predicting the sky as trees
(or buildings), or distorting other regions. Clearly, our proposed SSRC
loss outperforms benchmarking methods using only the SCC loss or only
the SRC loss. 

The effectiveness of our method is also demonstrated on the more
challenging Synthia dataset. Five refined images are shown in Figure
\ref{fig:syn2cs}. The refinement of Synthia to CityScapes is a
challenging task as it involves differences in viewpoints, classes, and
objects. For example, we have the top-down view in Synthia and the
ground view in CityScapes. More humans and objects exist in the Synthia
dataset. Our method can maintain realistic-looking images that resemble
those in the CityScapes dataset after refinement. 
 
\section{Conclusion}\label{sec:Conclusions}

A novel unsupervised synthetic image refinement method by combining
contrastive learning and semantic-structural relation consistency (SSRC)
was proposed in this work. By incorporating SSRC loss, semantic and
structural consistency between synthetic and refined images can be
maintained to reduces the semantic distortion effectively.  Consistency
in this context can be estimated by computing the mutual information in
both feature and pixel spaces. Furthermore, we exploited both the
semantic relation and hard negative mining to further improve the
performance of the contrastive loss function. Our method yields the
state-of-the-art performance. Its effectiveness was demonstrated
quantitatively and qualitatively. 

\section*{ACKNOWLEDGMENTS}       

This project was sponsored by US DoD LUCI (Laboratory University
Collaboration Initiative) fellowship and US Army Research Laboratory.
The authors also acknowledge the Center for Advanced Research Computing
(CARC) at the University of Southern California for providing computing
resources. The first author would also like to extend her gratitude to
Vasileios Magoulianitis for his valuable input in draft proofreading
and editing.

\bibliography{report} 
\bibliographystyle{spiebib} 

\end{document}